\documentclass[journal]{IEEEtran}

\usepackage{xcolor,soul,framed,caption}

\colorlet{shadecolor}{yellow}
\usepackage[pdftex]{graphicx}
\graphicspath{{../pdf/}{../jpeg/}}
\DeclareGraphicsExtensions{.pdf,.jpeg,.png}

\usepackage{threeparttable}
\usepackage{algorithm}
\usepackage{algpseudocode}
\usepackage{booktabs}
\usepackage{siunitx}
\usepackage{multirow}
\usepackage[cmex10]{amsmath}
\usepackage{amssymb}
\usepackage{bbm}
\usepackage{array}
\usepackage{mdwmath}
\usepackage{mdwtab}
\usepackage{eqparbox}
\usepackage{url}
\usepackage{hyperref}
\hypersetup{
    colorlinks=true,
    linkcolor=blue,
    filecolor=magenta,      
    urlcolor=cyan
}
    
\begin{document}
\bstctlcite{IEEEexample:BSTcontrol}
\title{Differentiable Integrated Motion Prediction and Planning with Learnable Cost Function for Autonomous Driving}

\author{Zhiyu Huang,
        Haochen Liu,
        Jingda Wu,
        and Chen Lv,~\IEEEmembership{Senior Member,~IEEE}

\thanks{Z. Huang, H. Liu, J. Wu, and C. Lv are with the School of Mechanical and Aerospace Engineering, Nanyang Technological University, Singapore, 639798. (E-mails: zhiyu001@e.ntu.edu.sg, haochen002@e.ntu.edu.sg, jingda001@e.ntu.edu.sg, lyuchen@ntu.edu.sg).}%
\thanks{This work was supported in part by the SUG-NAP Grant, Nanyang Technological University, and the A*STAR AME Young Individual Research Grant, Singapore (No. A2084c0156).}%
\thanks{Corresponding author: C. Lv.}}


\maketitle

\begin{abstract}
Predicting the future states of surrounding traffic participants and planning a safe, smooth, and socially compliant trajectory accordingly is crucial for autonomous vehicles. There are two major issues with the current autonomous driving system: the prediction module is often separated from the planning module and the cost function for planning is hard to specify and tune. To tackle these issues, we propose a differentiable integrated prediction-planning framework (DIPP) that can also learn the cost function from data. Specifically, our framework uses a differentiable nonlinear optimizer as the motion planner, which takes as input the predicted trajectories of surrounding agents given by the neural network and optimizes the trajectory for the autonomous vehicle, enabling all operations to be differentiable, including the cost function weights. The proposed framework is trained on a large-scale real-world driving dataset to imitate human driving trajectories in the entire driving scene and validated in both open-loop and closed-loop manners. The open-loop testing results reveal that the proposed method outperforms the baseline methods across a variety of metrics and delivers planning-centric prediction results, allowing the planning module to output trajectories close to those of human drivers. In closed-loop testing, the proposed method outperforms various baseline methods, showing the ability to handle complex urban driving scenarios and robustness against the distributional shift. Importantly, we find that joint training of planning and prediction modules achieves better performance than planning with a separate trained prediction module in both open-loop and closed-loop tests. Moreover, the ablation study indicates that the learnable components in the framework are essential to ensure planning stability and performance. Code and supplementary videos are available at {\href{https://mczhi.github.io/DIPP/}{https://mczhi.github.io/DIPP/}}.
\end{abstract}

\begin{IEEEkeywords}
Differentiable motion planning, multi-agent interactive prediction, autonomous driving
\end{IEEEkeywords}

\IEEEpeerreviewmaketitle

\section{Introduction}

\begin{figure}[t]
    \centering
    \includegraphics[width=\linewidth]{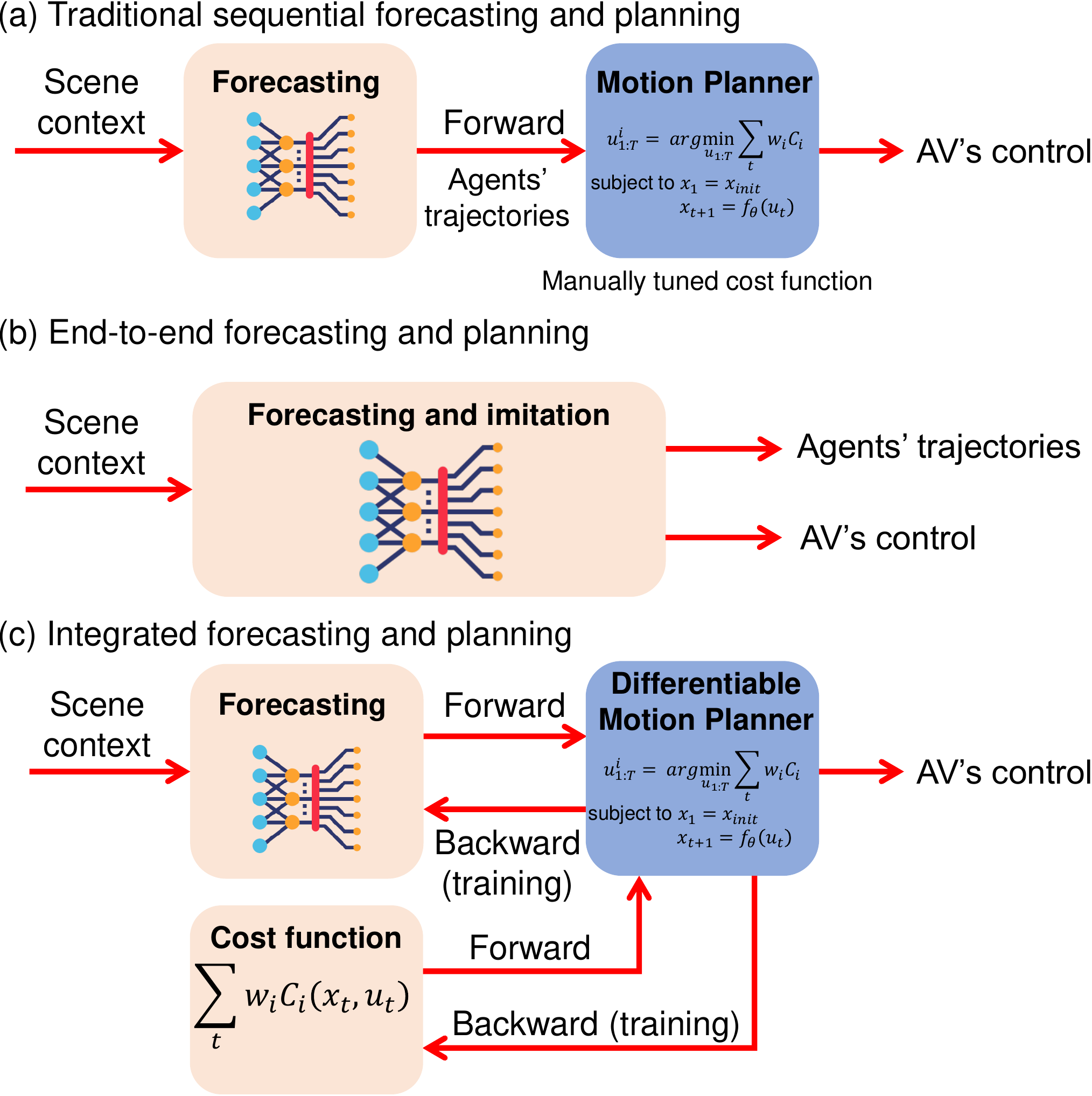}
    \caption{Three different motion planning paradigms: (a) traditional sequential prediction and planning; (b) end-to-end method; (c) our proposed method.}
    \label{fig1}
    \vspace{-0.55cm}
\end{figure}

\IEEEPARstart{M}{aking} safe, socially-compatible, and human-like decisions is a fundamental capability of autonomous vehicles (AVs). While learning-based end-to-end decision-making methods, such as imitation learning and reinforcement learning \cite{tampuu2020survey}, have enjoyed vigorous growth recently, they still lack interpretability, robustness, and safety compared to classic planning-based methods \cite{gonzalez2015review, ma2022local, hang}. Nonetheless, to make safe and smooth plans in complex traffic scenarios, the key is to accurately forecast the future trajectories of surrounding traffic participants \cite{huang2021multi, gao2021interacting, mo2022multi, chen2022heterogeneous}. The current autonomous driving stack treats prediction and planning as separate and sequential parts (see Fig. \ref{fig1}(a)), which means the prediction module is independent of the planning module. The problem with this setting is that prediction and planning tasks are highly interrelated and interdependent tasks, especially when the AV interacts with other traffic participants in urban areas. Moreover, another challenge with the traditional method is how to specify the cost function that can properly evaluate future motion plans and achieve a delicate trade-off between different requirements, e.g., collision avoidance, travel efficiency, and ride comfort. Manually tuning the parameters of the cost function is laborious and time-consuming, and might only be applicable in certain scenarios. Although some methods have been proposed to learn the cost function from driving data, such as continuous inverse optimal control (CIOC) \cite{levine2012continuous, xu2022energy}, sampling-based maximum-entropy inverse reinforcement learning (IRL) \cite{huang2021driving}, as well as max-margin method \cite{sadat2019jointly}, they are not coupled with the prediction module and rely on the impractical assumption that the prediction results are perfect. Some other methods have turned to a pure data-driven setting \cite{ngiam2021scene}, which utilizes a holistic model to directly output planned AV trajectory while predicting other agents' future trajectories (see Fig. \ref{fig1}(b)), implicitly handling the interactions between agents. However, such methods lack safety guarantees and reliability for the decision-making task in safety-critical applications.

In this paper, we propose a novel framework called differentiable integrated prediction and planning (DIPP), as shown in Fig. \ref{fig1}(c). The objective of the DIPP framework is to make the prediction module aware of the downstream planning task, so as to deliver planning-aware prediction results to the motion planner, without altering the established structure of the prediction-planning process. The principle of the DIPP framework is to implicitly change the prediction of other agents' future behaviors in the training process, where the planning error of the AV could also influence the prediction results of other agents, enabling the planner to make better and human-like plans at test time. Specifically, we construct a Transformer-based neural network to forecast the future trajectories of surrounding agents simultaneously. Then, the prediction results are channeled into a differentiable motion planner (optimizer), which explicitly plans a trajectory for the AV. When training the framework, the AV's planned trajectory is compared against the human driving one and the planning loss can be back-propagated to the prediction module and cost function, to alter the prediction results and automatically adjust the cost function with the objective of making human-like plans. Here, we assume the perception results (e.g., detecting and tracking of moving objects \cite{llamazares2020detection}, localization, and mapping) are accurate and focus on the joint prediction and planning problem. The proposed DIPP framework is trained to match the human driving trajectories (interactions among humans) in the entire driving scene and tested in both open-loop and closed-looped manners. In open-loop evaluations, we compared the planned trajectories of our framework directly against human trajectories without executing the plan, while in closed-loop evaluations, we unrolled the planned trajectory in simulation to test its performance in deployment scenarios. The results show that our framework is able to deliver more planning-centric prediction results, enabling the system better adapt to the scenarios in which they are deployed.

In summary, the main contributions of this paper are listed as follows:
\begin{enumerate}
\item We propose a fully differentiable structured learning framework that integrates prediction and planning modules for autonomous driving, enabling the prediction results better fit to the downstream planning task and the cost function learnable from real-world driving data.
\item We train the proposed framework with a large-scale real-world driving dataset that covers a wide variety of complex urban driving scenarios and validate it in both open-loop and closed-loop manners. 
\item We demonstrate that the proposed framework outperforms the baseline methods in both open-loop and closed-loop tests and conduct an ablation study to investigate the importance of each component.
\end{enumerate}
 
\section{Related Work}
\label{sec2}
\subsection{Multi-agent prediction}
There has been a growing body of learning-based approaches for trajectory prediction to excellent effect because deep neural networks can handle complex environments with multiple interacting agents and various road structures. Some works \cite{gao2020vectornet, varadarajan2021multipath++, huang2021multi, gu2021densetnt} utilizing vector representation of scene context and Transformer-based networks have advanced the forecasting accuracy even further. However, most of the existing models are formulated to predict each agent’s future trajectory independently, which is computationally inefficient and might produce inconsistent results. Thus, some approaches focus on multi-agent joint prediction to generate future trajectories for all agents of interest in a consistent manner \cite{gilles2021thomas, mo2022multi, zhou2022hivt, jia2021multiagent}. Importantly, using the multi-agent prediction method can allow the model to capture the interactions between agents and facilitate the planning task. Therefore, in our proposed framework, we employ the multi-agent prediction setting and provide each agent with a local vectorized map that shows the possible lanes and nearby crosswalks. We utilize Transformer modules to model the interactions between agents and their relations to different elements on the local map.

\subsection{Motion planning}
Motion planning is a long-researched area and there are a variety of approaches such as trajectory optimization, graph search, random sampling, and more recently learning-based methods. Learning-based methods employ neural networks as driving policies that generate actions or trajectories directly from sensor inputs or perception results, such as deep imitation learning \cite{chen2021data, huang2020multi} and reinforcement learning \cite{huang2021efficient, kendall2019learning, aradi2020survey}. Though simple and efficient, the neural network-based motion planner suffers from poor interpretability and generalization capability and also lacks stability and safety guarantees. Therefore, we turn to classic motion planning methods \cite{gonzalez2015review} and a popular choice is an optimization-based method as it is flexible and achieves maximal control of the trajectory. In particular, we employ a differentiable least-squares nonlinear optimizer as the motion planner, so that the optimizer can be integrated into the neural framework and its parameters (e.g., cost function weights) can be learned at the same time. In contrast to other planning methods that often work in static environments or assume a perfect prediction of surrounding obstacles, our proposed approach integrates the prediction of the surrounding agents and jointly trains the predictor and planner to achieve better performance of robustness.

\subsection{Joint prediction and planning}
Driving in dense and interactive traffic requires joint reasoning of other agents' future behaviors and the AV's plans \cite{cui2021lookout, espinoza2022deep, ngiam2021scene} to make active and human-like decisions. One promising framework is to generate a set of candidate trajectories for the AV and predict other agents' future trajectories conditioned on the AV's planned trajectories \cite{song2020pip}, and the best planning trajectory could be selected after evaluating these candidate plans and corresponding prediction results \cite{liu2021deep, cui2021lookout, casas2021mp3}. 
For example, an interactive prediction and planning framework is proposed in \cite{espinoza2022deep}, which can predict other agents’ future states according to a planned state sequence of the ego vehicle. The planner in their approach uses a cross-entropy method to sample a large number of action sequences and has to iteratively query the prediction model to get other agents’ responses, which may significantly slow down the planning process. In contrast, our method enables the prediction model to output planning-aware prediction results and implicitly captures the influence of the ego vehicle's future actions during the training process. This allows the planner to generate human-like plans in response to the prediction results during test time without requiring iterative queries to the prediction model.

Another promising framework is the end-to-end holistic neural network model that implicitly captures the prediction-planning interactions in the latent space, such as SceneTransformer \cite{ngiam2021scene}, which jointly outputs a motion plan for the AV and predicted trajectories for other agents. Although end-to-end models enjoy enhanced accuracy and simple inference, they cannot explicitly compute the feedback loop between planning and prediction from open-loop and offline training and thus cannot guarantee safety. Another challenge of such methods is the distribution shift encountered when the predicted actions are rolled out in deployment. Therefore, some other methods, e.g., MATS (Mixtures of Affine Time-varying Systems) \cite{ivanovic2020mats}, have tried to couple the prediction model with classic optimization methods because of their interpretability and reliability. MATS predicts the parameters of a linear-affine dynamical structure, which are utilized by a model predictive controller (MPC) for motion planning. Though efficient due to a small number of parameters, MATS does not incorporate the scene context in forecasting, and its performance is compromised. In this paper, we propose to output the trajectories of surrounding agents using a neural network and combine it with differentiable optimization steps that explicitly consider ride comfort, safety, and traffic rules, making the feedback between planning and prediction differentiable and enhancing both safety and human driving similarity.


\section{Method}
\subsection{Problem formulation}
We formulate the driving scene with the AV and a varying number of diverse interacting traffic participants as a continuous-space discrete-time system (time discretization is uniform), where the AV is denoted as $A_0$ and other agents are denoted as $A_1, \dots, A_{N}$. Each agent including the AV has a semantic class (i.e., vehicle, bicycle, or pedestrian), and its state at time $t$ is denoted as $\mathbf{s}_t^i$, where $i$ is the agent index. We also introduce the scene context, such as a vectorized high-definition map and traffic light signals, to the system and denote it as $\mathbf{M}$. Assuming the current time step is $t=0$, given the historical states of all agents for the previous $H$ timesteps $\mathbf{X} = \mathbf{s}^{0:N}_{-H:0}$ and the scene context $\mathbf{M}$, the predictor needs to predict $K$ possible joint future trajectories of all agents for the next $T$ timesteps $\{\mathbf{Y}_k|k=1,\cdots,K\}$, $\mathbf{Y}_k = \mathbf{\hat s}^{0:N}_{1:T}$ where $\mathbf{\hat s}_t^i$ is the predicted state of agent $i$ at future timestep $t$, and the probability of each future $\{p_k|k=1,\cdots,K\}$. The planner needs to further optimize the AV's future trajectory according to the initial trajectory of the AV $\mathbf{\hat s}^{0}_{1:T}$, the prediction results of other agents, and the cost function.

\begin{figure*}[htp]
    \centering
    \includegraphics[width=0.95\linewidth]{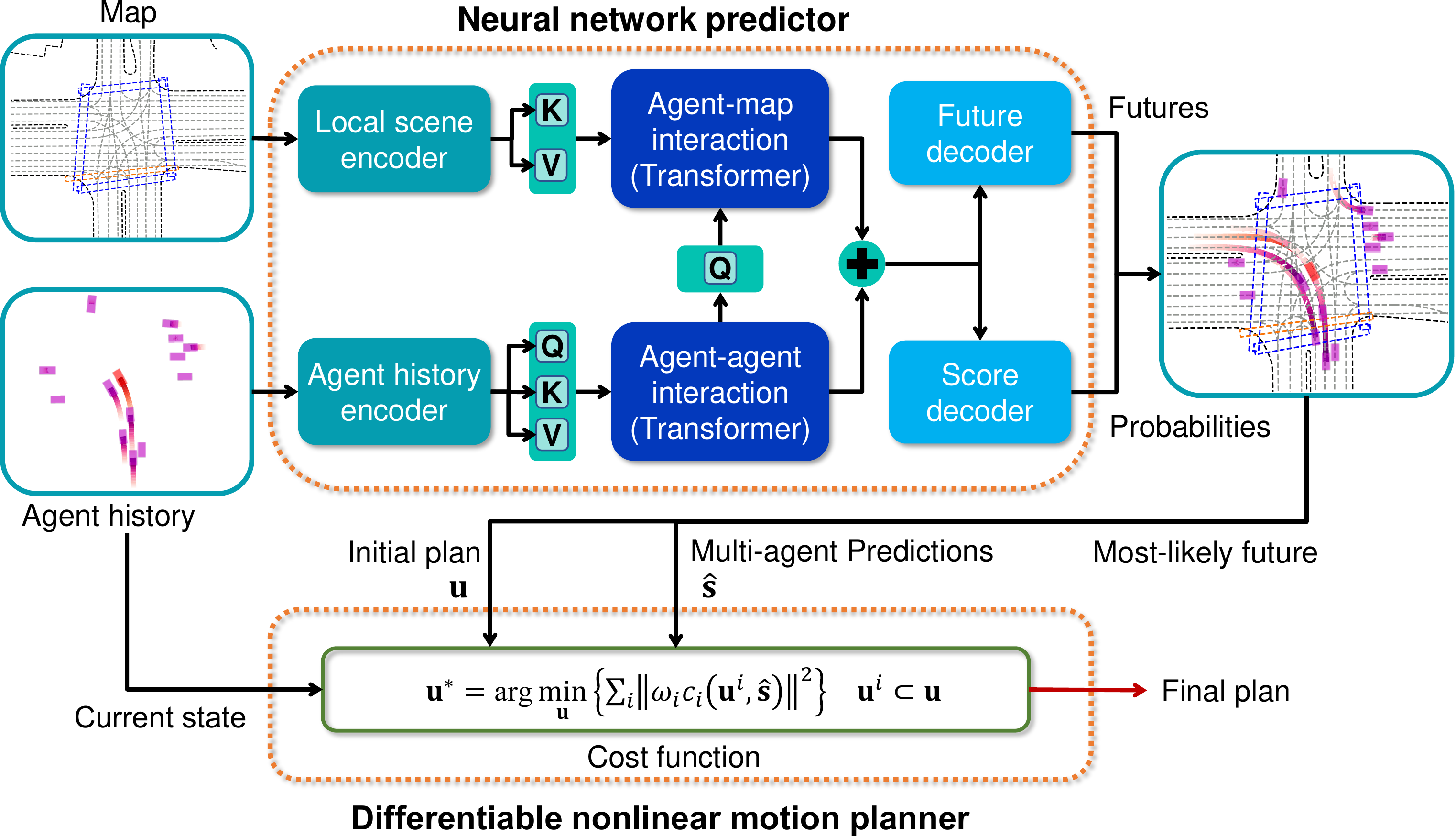}
    \caption{The proposed differentiable integrated prediction and planning framework. The neural network predictor is utilized to predict the future states of surrounding agents and the initial plan for the motion planner, and the differentiable motion planner with a learnable cost function is employed to explicitly plan an AV trajectory. All the components are connected and end-to-end differentiable.}
    \label{fig2}
    \vspace{-0.5cm}
\end{figure*}

Accordingly, the proposed framework consists of two parts, as shown in Fig. \ref{fig2}. First, we build up a neural network predictor to embed the historical trajectories of agents and scene context into high-dimensional spaces. The hidden embeddings are used to generate the Keys, Values, and Queries ($\mathbf{K}$, $\mathbf{V}$, and $\mathbf{Q}$) used in the attention mechanism \cite{vaswani2017attention} of the interaction modeling modules (i.e., agent-map interaction and agent-agent interaction). All agents' different future trajectories and their probabilities are finally decoded from the latent representation of the interaction relationships. Second, we employ a differentiable optimizer as a motion planner to explicitly plan a future trajectory for the AV according to the most-likely prediction result and initial motion plan. Specifically, we formulate the motion planning problem as a nonlinear least squares problem, where $\mathbf{u}$ is the optimization variables (action sequence), $\mathbf{u}^i \subset \mathbf{u}$ is a subset of the action sequence, $\mathbf{\hat s}$ is the predicted states of other agents; $c_i$ is the individual cost, and  $\omega_i$ is the weight of the cost term. The optimization objective (sum of squared cost terms) covers a variety of factors in driving, including travel efficiency, collision avoidance, and ride comfort. We employ a differentiable nonlinear optimizer to solve the optimization, and thus the gradient from the planner (planning loss) can be back-propagated to the prediction module and cost function in the training stage, making the whole framework end-to-end learnable. The details of the neural network predictor and differentiable motion planner are given in Section \ref{nn} and Section \ref{mp}, respectively.

\subsection{Multi-agent prediction and initial plan}
\label{nn}
\textbf{Input representation}. The neural network predictor receives two types of input data: historical agent states and scene context. For each agent, its historical state is a sequence of dynamic features for the past timesteps $H=20$, including its 2D position, heading angle, velocity, and bounding box size. We consider the nearest $N=10$ agents around the AV, whose observation data is stored in a fixed-size tensor with the missing agents padded as zeros. For the scene context, we consider two types of map elements, which are lanes and crosswalks. Each map element is presented by a vector, which consists of a sequence of waypoints with different associated features. For each agent, we build a local scene context, which encompasses the possible lanes it might take and surrounding crosswalks. The features of the waypoint in the drivable lane include the positions and headings of the lane center and lane boundaries, as well as the speed limit and traffic signals (e.g., traffic light state and stop sign), and the features of the waypoint in the crosswalk are the positions of points enclosing the crosswalk area. All the agents' and map elements' positional attributes are translated to the AV's local coordinate system.

\textbf{Encoding agent history and scene context}. To encode an agent’s observed historical states, we feed the state sequence into a long short-term memory (LSTM) network. LSTM is used because we find it would bring better final prediction performance and computational efficiency than Transformers when dealing with short-term time series, and we can retrieve the encoded state feature at the last timestep from the LSTM as the node feature in the agent interaction graphs. Each type of agent shares an LSTM encoder, and all agents including the AV are stacked to form a tensor of agents' encoded historical states. The local scene context encoder consists of a lane encoder for processing the lanes and a crosswalk encoder for processing the crosswalk vectors. The lane encoder uses a multi-layer perceptron (MLP) to encode numeric features (e.g., positions, directions, and speed limits) and embedding layers to encode discrete features (e.g., traffic light state, lane type, and stop sign), and the crosswalk encoder is another MLP to encode the numeric features (i.e., positions and headings). For each agent, we find its nearby 6 lanes and 4 crosswalks as map vectors, encode them separately, and concatenate the encoded map vectors to form a tensor of the agent's local scene context.

\textbf{Modeling agent-agent and agent-scene interactions}. Capturing relationships or dependencies between agents and the environment is essential in ensuring prediction accuracy. To realize this, we follow the idea of \cite{mo2022multi} and construct two interaction graphs, which are the agent-agent interaction graph (all agents are fully connected in the graph) and the agent-map interaction graph (a target agent is connected to all local lanes and crosswalks polylines. We use a two-layer self-attention Transformer encoder as the agent-agent interaction encoder to process the graph, where the query, key, and value ($\mathbf{Q}, \mathbf{K}, \mathbf{V}$) are the encoded agents' historical trajectory features. Here, the functionality of the Transformer is similar to the graph attention network \cite{mo2022multi} but with improved structure and representation capability, and using a few layers of Transformer blocks would not significantly increase the computational cost. With each agent's encoded local scene context in hand, we employ two cross-attention Transformer encoders as the agent-map interaction encoder: one is for modeling the agent's attention on each lane vector (focusing on waypoints in the vector), and the other is for agent's attention on each crosswalk vector. We utilize an agent's interaction feature as query ($\mathbf{Q}$), and a single map vector (a sequence of encoded waypoints) as the key and value ($\mathbf{K}, \mathbf{V}$). The operation is called multiple times to process all the map vectors from an agent, resulting in a sequence of agent-map vectors attention features. Then we introduce a multi-modal attention Transformer encoder \cite{huang2021multi}, which is essentially an ensemble of cross-attention modules, to output multiple trajectories through the different attention modules, representing the different relations between the agent and map vectors. We still use an agent's interaction feature as query, and agent-map vectors attention as key and value, yielding different encodings of the agent's relationship with the local scene context. Likewise, we apply the multi-modal encoder to all agents to extract their possible interactions with the scene context and stack the results along the future axis.

\textbf{Decoding multi-modal future}. For each agent, its agent-agent interaction encodings are repeated and concatenated with the multi-modal agent-map interaction encoding to form a final feature vector. For the surrounding agents, we decode their possible future trajectories through a shared MLP from the final feature vector. For the AV, we decode its future control actions (i.e., acceleration and steering angle) from the feature vector through an MLP and translate the action sequence to trajectory using a kinematic model (see Section \ref{mp}). We also output multiple possible trajectories for the AV to better model the interaction between agents. To predict the probability of each future (joint trajectories for all agents), we use max-pooling to aggregate the information from all agents and map vectors and an MLP to decode the probability. The motion planner will take as input the future (i.e., the initial plan and other agent's prediction) with the highest probability. 

\subsection{Differentiable motion planning}
\label{mp}
\subsubsection{Problem statement}
We consider an open-loop finite-horizon optimal planning problem, which is to find a sequence of control inputs $\mathbf{u} = \{\mathbf{u}_1, \mathbf{u}_{2}, \cdots, \mathbf{u}_{T} \}$ to minimize the cost function in Eq. \ref{eq1}. The cost function is a sum of squared residual terms, each represented by a product of a weight $\omega_i$ and cost $c_i$ \cite{torch}, and the state of other surrounding agents $\mathbf{\hat s}$ is provided by the neural network predictor.
\begin{equation}
\label{eq1}
\mathbf{u}^{*} = \arg \min_{\mathbf{u}} \frac{1}{2} \sum_i \parallel {\omega_i c_i(\mathbf{u}^i, \mathbf{\hat s})} \parallel^2,    
\end{equation}
where $c_i$ is a function of subsets of the control sequence $\mathbf{u}^i \subset \mathbf{u}$ and potentially the predicted states of other surrounding agents $\mathbf{\hat s}$. In the training process, the gradient of the final planning error can be backpropagated to the predicted states of other agents and cost weights through the optimization problem \cite{amos2018differentiable, bhardwaj2020differentiable}, to implicitly influence other agents' future behaviors and adjust the cost weights.

\subsubsection{Kinematic model}
When calculating some costs, we need to convert the control action $\mathbf{u}_t = \{ a_t, \delta_t \}$ (where $a_t$ is acceleration and $\delta_t$ is steering angle) to state $\mathbf{s}_t = \{x_t, y_t, \theta_t, v_t\} $ (where $(x_t, y_t)$ is 2D position, $\theta_t$ is heading angle, and $v_t$ is velocity). We adopt the kinematic bicycle model \cite{rajamani2012lateral} shown in Eq. \ref{eq2} to update the states of the AV.
\begin{equation}
 \begin{aligned} 
\label{eq2}
v_{t+1} &= v_t + a_t \Delta t, \\
x_{t+1} &= x_t + v_t \cos(\theta_t) \Delta t, \\
y_{t+1} &= y_t + v_t \sin(\theta_t) \Delta t, \\
\theta_{t+1} &= \theta_t + \frac{v_t}{L} \tan(\delta_t) \Delta t,
\end{aligned} 
\end{equation}
where $L$ is the wheelbase of the vehicle and $\Delta t$ is the time interval. The kinematic bicycle model is differentiable, thus permitting calculating gradients and Jacobians in the optimizer.

\subsubsection{Cost function} 
The cost function encompasses a variety of carefully crafted cost terms that encode different aspects of driving including travel efficiency, ride comfort, traffic rules, and most importantly safety. The details of the different cost terms are given below.

\textbf{Travel efficiency}. 
We encourage the AV to reach the destination as fast as possible but not run above the speed limit. Therefore, the cost of travel efficiency is defined as the difference between the AV's speed $v_t$ and speed limit:
\begin{equation}
\mathbf{c}_t^{speed} = v_t - v_{limit},    
\end{equation}
where $v_{limit}$ is the speed limit of the lane.

\textbf{Ride comfort}. 
Human drivers prefer comfortable maneuvers, and we introduce four sub-costs to represent the ride comfort factors the AV needs to optimize. They are longitudinal acceleration $a_t$ and longitudinal jerk $\dot a_t$, as well as steering angle $\delta_t$ and steering change rate $\dot \delta_t$ for lateral stability and comfort.
\begin{equation}
 \begin{aligned} 
\label{eq4}
\mathbf{c}_t^{acc} = a_t, \\
\mathbf{c}_t^{jerk} = \dot a_t, \\
\mathbf{c}_t^{steer} = \delta_t, \\
\mathbf{c}_t^{rate} = \dot \delta_t,
\end{aligned} 
\end{equation}
where $\dot a_t = \frac{\Delta a_t}{\Delta t}$ and $\dot \delta_t = \frac{\Delta \delta_t}{\Delta t}$ are the discrete forms of the longitudinal jerk and steering rate respectively.

\textbf{Traffic rules}. 
The AV is expected to adhere to the traffic rules and structure of the road. Thus, to promote staying close to the lane (route) centerline and following the lane direction, two sub-costs are designed.
\begin{equation}
\begin{aligned} 
\label{eq5}
\mathbf{c}_t^{pos} = p_t - p_{l, \perp}, \\
\mathbf{c}_t^{head} = \theta_t - \theta_{l, \perp},
\end{aligned} 
\end{equation}
where $p_t$ and $p_{l, \perp}$ are the positions of the AV and closest point from the lane's centerline to the AV respectively, and $\theta_t$ and $\theta_{l, \perp}$ are the heading angles of the AV and its closest point on the lane respectively.

In addition, obeying traffic lights should be treated as a hard constraint for the AV. Here, we replace the hard constraint with a soft penalty term, which can be assigned with a large cost weight. We assume the AV runs along a predefined route and its running distance is $s_t$, which is derived from $s_t=\sum_{t^{\prime}=1}^t v_{t^{\prime}} \Delta t$, and the stop line position (if encountering a red light) on the route is $s_{stop}$. We can formulate the cost of violating traffic lights using the hinge loss:
\begin{equation}
\label{eq6}
\mathbf{c}_t^{traffic} = \begin{cases} s_t - s_{stop}, & s_t \geq s_{stop} \\ 0, & \text{otherwise} \end{cases}.
\end{equation}
It means that if the AV goes past the stop line at a red light, a large penalty will be induced, thus forcing the AV to stop near the stop point.

\begin{figure}[htp]
    \centering
    \includegraphics[width=0.95\linewidth]{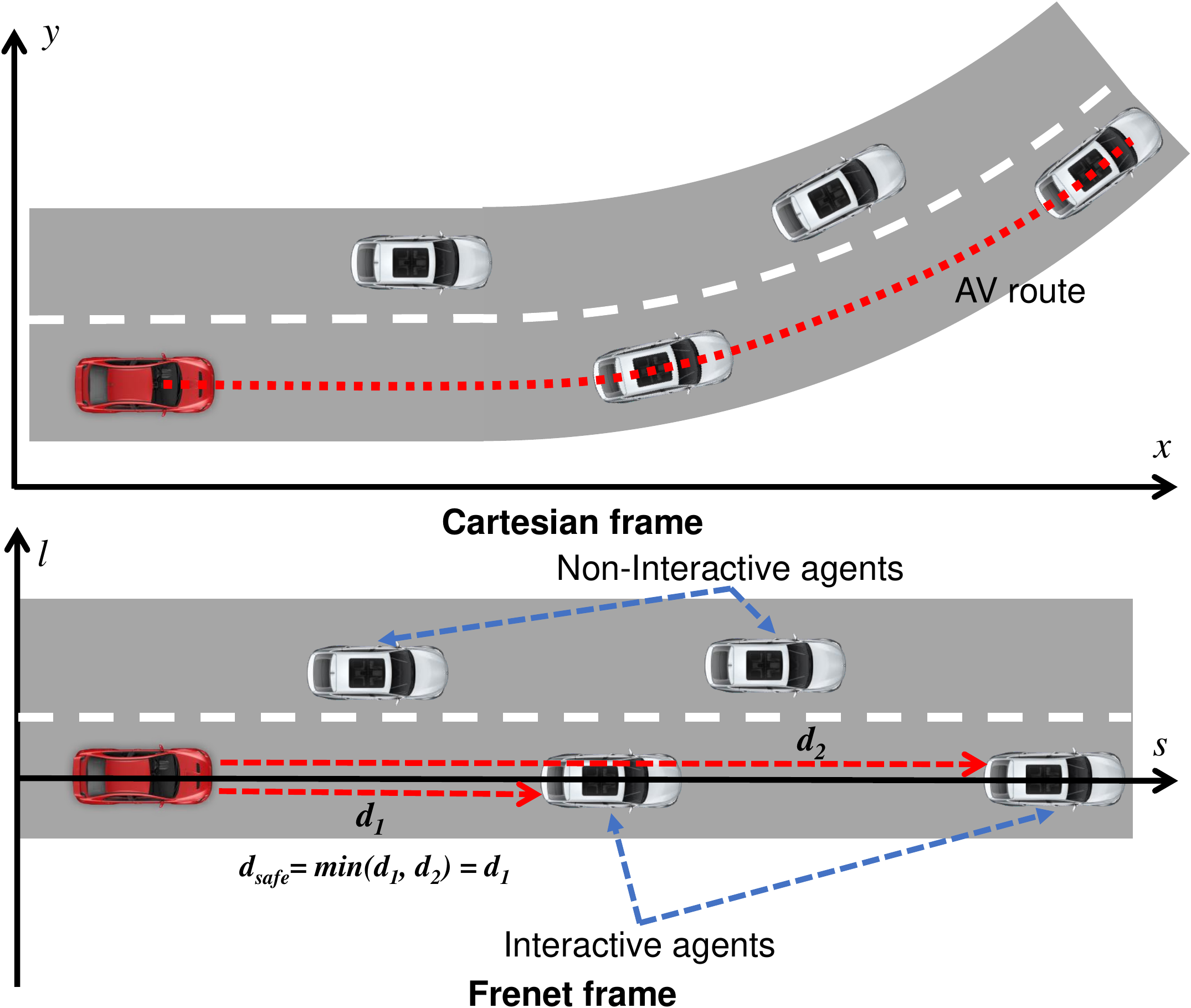}
    \caption{Illustration of the calculation of safe distance. Other agents are first projected to the Frenet frame to find the interacting agents, and the calculation of distances is in the Cartesian frame.}
    \label{fig3}
\end{figure}

\textbf{Safety}. 
Keeping a safe distance from other traffic participants on the road to avoid collision is a fundamental requirement for AVs. However, optimizing the safe distances to all other agents in the scenario is unnecessary and time-consuming. Therefore, we take advantage of the Frenet frame\cite{zhang2019optimal, zhang2020optimal}, which decouples the vehicle trajectory into the longitudinal direction along a predefined driving route and the lateral direction perpendicular to the route, to define the interactive agents. As illustrated in Fig. \ref{fig3}, all other agents' positions are projected to the Frenet frame with the AV's route as the reference path. At each future timestep, those agents whose predicted positions are within the route's conflict area are defined as the interactive agents. The calculation of safe distance at a specific timestep is given as:

\begin{equation}
\label{eq7}
d_{safe} = \min_{i} \parallel p_t - p_{t}^i \parallel_2,
\end{equation}
where $p_{t}^i$ is the predicted position of the interactive agent $i$ at future timestep $t$.

We also employ the hinge loss to formulate the safety cost to ensure the safe distance is large enough.
\begin{equation}
\label{eq8}
\mathbf{c}_t^{safety} = \begin{cases} \epsilon - d_{safe}, & d_{safe} \leq \epsilon \\ 0, & \text{otherwise} \end{cases},
\end{equation}
where $\epsilon$ is the minimum safety distance requirement, which is defined as the sum of the lengths of two agents and a safety gap.

\textbf{Weights}. 
The cost terms in Eq. \ref{eq1} consist of different features listed before at different timesteps. Here, we only consider the weights for different features and the weights for individual features are the same at different timesteps. For features other than safety and traffic rules, we consider all the timesteps over the planning horizon. For the cost of traffic rule violation, we consider every other timestep across the future horizon to reduce computation costs. Because the computation for the safety cost is expensive, we only consider the trajectory waypoints at $[0.1, 0.3, 0.6, 1.0, 1.5, 2.0, 2.5, 3.0, 4.0, 5.0]$ seconds, which could significantly speed up the optimization without compromising safety. To ensure the constraints are satisfied, the weights for the constraints (i.e., safety and traffic light violations) in the cost function, are set to a large enough value and fixed during training.

\subsubsection{Differentiable optimization} 
The Gauss-Newton algorithm is utilized to solve the nonlinear optimization problem \cite{bhardwaj2020differentiable}. It is an iterative least-squares optimization approach, where at each iteration step, the nonlinear objective is first linearized around the current control variables to derive the linear system:
\begin{equation}
\label{eq10}
\left( \sum_i J_i^{\top} J_i \right) \Delta \mathbf{u} = \sum_i J_i^{\top} \omega_i c_i,
\end{equation}
where $J_i = \frac{\partial (\omega_i c_i)}{\partial \mathbf{u}^i}$ is the Jacobian of the cost with
respect to the control variable.

We can explicitly solve the linear system using Cholesky decomposition of the normal equations to find the update $\Delta \mathbf{u}$, and eventually update the control variables:
\begin{equation}
\label{update}
\mathbf{u} \leftarrow \mathbf{u} - \alpha \Delta \mathbf{u}.
\end{equation}
where $0 < \alpha \leq 1$ is the step size. In the original Gauss-Newton method, $\alpha=1$, but it may cause the nonlinear objective not to decrease at every iteration. Therefore, an improved version is setting $0 < \alpha < 1$ to employ a fraction of the update and thus mitigates the divergence issue.

Since the above procedure is fully differentiable, we can set up other differentiable components (e.g., prediction and cost function weights) and integrate them into the planner, enabling the whole architecture to be differentiable. Moreover, to start the optimization, one has to provide an initial guess for the control variables, which is crucial to the convergence to the global optimum, and we use the control actions given by the prediction network as the initial guess, which is also differentiable.  

\subsection{Learning process}
At a high level, we regard the motion planner as the primary component of our framework. During the forward pass, the motion planner takes as input the prediction results of other agents and the initial plan (with the highest probability), as well as the cost function, to start the trajectory optimization and iteratively updates the planned trajectory until passing a convergence check or reaching the maximum number of iterations. At the end of the optimization, the motion planner will output a trajectory for the AV to follow. In the backward pass, we evaluate the loss function on the planned trajectory, compute gradients with respect to the initial plan, cost function weights, as well as the predicted trajectories of other agents, and back-propagate the loss through the neural components and update the parameters, in order to generate better quality trajectories. The learning process of our framework is stable and efficient because all the components (i.e., prediction, initial plan, and cost function) are learned in a supervised manner. Compared to generative adversarial imitation learning (GAIL) \cite{ho2016generative, baram2017end}, which learns proxy reward signals through adversarial learning, our method only learns the weights of the cost function terms. However, the initialization of the cost function weights is important in our method because this can affect the optimization process. The learning process of our framework is summarized in Algorithm \ref{alg1}.

Since all operations in the framework are differentiable, we can train the entire framework in an end-to-end fashion from real-world driving data, and we implement the framework using PyTorch and Theseus \cite{torch}. The loss function for training the framework encompasses four terms: prediction loss for all agents, score loss to accurately predict the probabilities of different futures, as well as imitation loss and planning cost for the AV. The overall loss is defined as:
\begin{equation}
\label{loss}
\mathcal{L} = \lambda_1 \mathcal{L}_{prediction} + \lambda_2 \mathcal{L}_{score} + \lambda_3 \mathcal{L}_{imitation} + \lambda_4 \mathcal{L}_{cost},
\end{equation}
where $\lambda_1$, $\lambda_2$, $\lambda_3$, and $\lambda_4$ are the weights to scale these different loss terms. For prediction loss, we treat each future to be coherent individual futures across all agents and back-propagate the smooth L1 losses on the trajectories through the joint future that most closely matches the ground truth:
\begin{equation}
\mathcal{L}_{prediction} = \sum_{k=1}^K \mathbbm{1}(k=\hat k) \sum_{i=1}^N \text{smooth}{L_1} (\xi^k_i - \xi_i^{gt}),
\end{equation}
where $\mathbbm{1}$ is the indicator function, $\hat k$ is the index of the best-predicted joint future of multiple agents (the future with the smallest sum of displacement errors across all agents, $\hat k = \arg \min_k \sum_i \parallel \xi^k_i - \xi_i^{gt}\parallel_2$), and $\xi_i^{gt}$ is the individual ground-truth trajectory. The scoring loss is defined as:
\begin{equation}
\mathcal{L}_{score} = \sum_{k=1}^K \mathbbm{1}(k=\hat k) \log p_k, 
\end{equation}
where $p_k$ is the probability of future $k$.

The motion planner would take as input the most-likely trajectories of all agents, where the AV's trajectory (original control actions) is used as planning initialization and other agents' future trajectories are used in computing the safety cost. The output of the motion planner $\xi_{AV}$ is compared against the ground-truth trajectory $\xi_{AV}^{gt}$ to calculate the imitation loss.
\begin{equation}
\mathcal{L}_{imitation} = \text{smooth}{L_1} (\xi_{AV} - \xi_{AV}^{gt}).  
\end{equation}

\begin{algorithm}
\caption{Learning of the proposed DIPP framework}
\begin{algorithmic}[1]
\Require Neural network predictor $f_{\theta}$, initial cost function weights $\{\omega_i\}$, differentiable planner $g$
\State Get the environment information (Agent tracks $\mathbf{X}$ and vectorized map $\mathbf{M}$)
\State Predict the AV and other agents' multi-modal joint trajectories $\{\xi^{k}_{i}\}_{k=1:K, i=0:N}, \ \{p_k\}_{k=1:K} \gets f_{\theta} \left(\mathbf{X}, \mathbf{M} \right)$
\State Calculate prediction loss $\mathcal{L}_{prediction}$ and score loss $\mathcal{L}_{score}$
\State Get the most-likely prediction result $\{\xi^{\hat k}_{i}\}_{i=0:N}, \ \hat k = \arg \min_k \sum_i \parallel \xi^k_i - \xi_i^{gt}\parallel_2$ \algorithmiccomment{Testing $\hat k = \arg \max \{p_k\}$}
\State Plan the AV's trajectory using the differentiable planner $\xi_{AV}, \mathcal{L}_{cost} \gets g \left( \xi^{\hat k}_{0}, \{\xi^{\hat k}_{i}\}_{i=1:N}, \{\omega_i\} \right)$
\State Calculate imitation loss $\mathcal{L}_{imitation}$
\State Calculate total loss $\mathcal{L}$ according to Eq. (\ref{loss})
\State Backpropagate loss and calculate gradients with respect to $\theta$ and $\{\omega_i\}$
\State Update $f_{\theta}$ and $\{\omega_i\}$ using Adam optimizer
\end{algorithmic}
\label{alg1}
\end{algorithm}

\section{Experiments}
\subsection{Dataset}
We train and validate our framework on the Waymo Open Motion Dataset \cite{ettinger2021large}, a large-scale real-world driving dataset that focuses on urban driving scenarios with complex and diverse road structures and traffic interactions. The dataset contains \num{103354} unique driving scenes, each 20 seconds in duration, with detailed map data and agent tracks. We randomly select 10\% of the dataset, which corresponds to 100 data files out of 1000 files in total, resulting in a total of \num{10156} different driving scenes. Since the driving scenes are randomly stored in the data files and the selected driving scenes are a random subset of the dataset, the data distribution of the reduced dataset remains the same as the original. To segment a 20-second driving scene into frames (each frame contains 7-second object tracks at 10Hz), we set up a 7 s time window with a 2 s history observation horizon and a 5 s planning/prediction horizon. The window slides from the beginning of the scene with a stride of 10 timesteps (1 second), resulting in 14 segments of 7 s frames. For training, we use 80\% of the selected scenes, while the remaining 20\% are used for validation and open-loop testing. The total number of frames used for training is \num{113622}, while \num{28396} frames are used for validation and open-loop testing.

\subsection{Evaluation}
We evaluate the performance of the framework in both open-loop and closed-loop manners. In the open-loop testing, the motion planner plans a trajectory for the AV based on the current states and prediction results at each frame, and we compare the planned trajectory against the AV's ground-truth trajectory. In the closed-loop testing, we construct a log-replay simulator, where at each step, the AV will take the first action from its planned trajectory and update its state, while the neighboring agents will follow their record trajectories in the dataset. We set up the following metrics to evaluate the performance of the framework.

\textbf{Safety}: the collision rate is employed to measure the safety performance of the system. For open-loop testing, collision is calculated based on the AV's planned trajectory and the ground-truth future trajectories of other agents. If a collision is detected at any step in the plan, then the frame is deemed as a collision. For closed-loop testing, we check if the AV collides with other agents at every time step during the simulation.

\textbf{Traffic rule violation}: we consider deviating from the route and passing over the stop line at a right light as traffic rule violations, and we calculate the total number of frames where the AV violates the traffic rules.

\textbf{Vehicle dynamics}: three metrics are introduced to quantify rider comfort and feasibility of the planned trajectory, which are longitudinal acceleration and jerk, as well as lateral acceleration. They are absolute values averaged over time in a scene and will be compared against human driving metrics.

\textbf{Human driving similarity}: we use the position errors between the planned trajectory in open-loop testing or rollout trajectory in closed-loop testing and the ground truth to quantify the human likeness. We calculate the position errors at different timesteps (i.e., 1 s, 3 s, 5 s in open-loop testing and 3 s, 5 s, 10 s in closed-loop testing).

\textbf{Prediction}: we calculate the average displacement error (ADE) and final displacement error (FDE) of the multi-agent joint trajectories from the most-likely future to reflect the performance of the prediction module. ADE computes the L2 norm between the ground-truth state $s_{a, t}$ and the most-likely joint prediction $\hat s^{\hat k}_{a, t}$: $\frac{1}{AT} \sum_a \sum_t \parallel s_{a, t} - \hat s^{\hat k}_{a, t} \parallel_2$, where $\hat k = \arg \max p_k $ is the index of the most-likely predicted joint future. FDE calculates the L2 distance between the ground-truth and predicted states at the final step $T$: $\frac{1}{A} \sum_a \parallel \hat s_{a, T} - \hat s^{\hat k}_{a, T}\parallel_2$.

\subsection{Comparison baselines}
We compare our proposed method with the following baselines to reveal the effects and advantages of the proposed framework.

\textbf{Vanilla imitation learning}: based on the same scene context, we use only imitation learning (IL) to train a network to directly generate the trajectory for the AV. The structure of the network is the same as the proposed method, but the prediction of other agents is removed and only one trajectory instead of multi-modal trajectories is generated. The IL policy is trained on the same dataset with the same hyperparameters as the proposed method, which makes sure the comparison results are fair.

\textbf{Imitation learning with prediction sub-task}: we train a multi-task neural network with the main task of imitation learning and the sub-task of predicting other agents' future trajectories. This is equivalent to the proposed network without the differentiable motion planner, and the training settings and hyperparameters are the same as the proposed method.

\textbf{Separated prediction-planning}: based on the same network, we separately train a prediction module without the integrated motion planner to output the initial trajectory for the AV and predicted trajectories for other agents, which is the same as the IL with the prediction model. In testing, a motion planner, which is the same as the proposed method, is utilized to perform trajectory planning for the AV with the trained prediction model and a pre-specified cost function. 

\textbf{Conservative Q-learning (CQL)} \cite{kumar2020conservative}: we utilize a widely-used offline reinforcement learning algorithm that can learn to make decisions from offline datasets. We implement the CQL method with the d3rlpy offline RL library \cite{d3rlpy} using the default hyperparameters. The observation space is a 3-channel RGB bird-eye-view image $256\times256\times3$ of the driving scene \cite{huang2022recoat}, the action is the target pose of the next step $\left(\Delta x, \Delta y, \Delta \theta \right)$ relative to the ego vehicle's current position, and the reward function is the distance traveled per step plus an extra reward for reaching the goal. The policy is trained on the same scenarios as the proposed method and all transition steps (observations, actions, rewards, and terminals) in each scenario will be used. The trained policy is only evaluated in the closed-loop planning test.

\textbf{Intelligent driver model (IDM)} \cite{zhang2022bayesian}: we utilize a well-known model-based driving model, IDM, to control the longitudinal movement of the ego vehicle along the reference route. The IDM planner takes as input the current state of the ego vehicle (position and speed) and the state of the leading agent (position, speed, and length) under the coordinate of the reference route, and outputs the acceleration of the ego, which is then translated to the coordinate in the scene. The IDM method is only used in the closed-loop planning test.

\subsection{Implementation details}
\subsubsection{Network}
The network outputs $3$ possible futures (joint trajectories of all agents) and their probabilities. For other agents, we predict the displacements relative to their current locations instead of their coordinates, which shows significant improvement in prediction accuracy. The learning of the cost function is also incorporated into the network and implemented by a small MLP, which takes a fixed dummy input and outputs the weights of the cost function. We set the cost weights for traffic light violations and collisions as large values (larger than two orders of magnitude) and unlearnable to ensure the constraints are satisfied. 

\subsubsection{Training}
We use a batch size of 32 and an Adam optimizer with a learning rate that starts from 2e-4 and decays by a factor of 0.5 every 4 epochs. The total number of training epochs is 20 and we pre-train the framework for 5 epochs without the planner to get good initial prediction results and control actions. For speed considerations, we set the max number of iterations for the motion planner to 2, which will also encourage the network to produce high-quality initial plans, and the step size for the Gauss-Newton update is $\alpha = 0.4$. The weights for the loss function are set to $\lambda_1 = 0.5$, $\lambda_2 = 1$, $\lambda_3 = 1$, and $\lambda_4 = 0.001$. The results shown in Sections \ref{r-open} and \ref{r-closed} are based on this loss function setting and we investigate the effects of the loss function weights in Section \ref{abla}. We clip the gradient norm of the network parameters with the max norm of the gradients as 5.

\subsubsection{Testing}
In the testing process, the max number of iterations for the motion planner is set to 50, the step size is set to $0.2$, and the absolute error tolerance is set as 1e-2, in order to plan a high-quality trajectory for the AV. Acceleration and jerk in the longitudinal and lateral directions are calculated based on the positions and headings on the trajectory. To check if a collision happens, we use a list of circles to approximate an object at each timestep. If the distance between any two circles' origins of the given two objects is lower than a threshold, it is considered as they collide with each other. 

The main hyperparameters used in this paper are summarized in Table \ref{tab0}.

\begin{table}[htp]
\centering
\caption{Hyperparameters of the model and training process}
\begin{tabular}{@{}ll@{}}
\toprule
Parameter                               & Value \\ \midrule
Number of neighbors  $N$                & 10      \\
Number of predicted futures $K$         & 3      \\
Historical timesteps $H$                & 20       \\
Future timesteps $T$                    & 50      \\  \midrule
Step size $\alpha$                      & 0.4      \\
Planning iterations                     &  2   \\
Initial learning rate                   & 2e-4     \\
Total training epochs                   &  20     \\
Pretraining epochs                      & 5      \\
Batch size                              & 32       \\
Weight for prediction loss $\lambda_1$  & 0.5      \\
Weight for score loss $\lambda_2$       & 1      \\
Weight for imitation loss $\lambda_3$   & 1      \\ 
Weight for planning cost $\lambda_4$    & 1e-3      \\\bottomrule
\end{tabular}
\label{tab0}
\end{table}

\section{Results and discussions}
\subsection{Open-loop test}
\label{r-open}

\begin{figure*}[htp]
    \centering
    \includegraphics[width=0.93\linewidth]{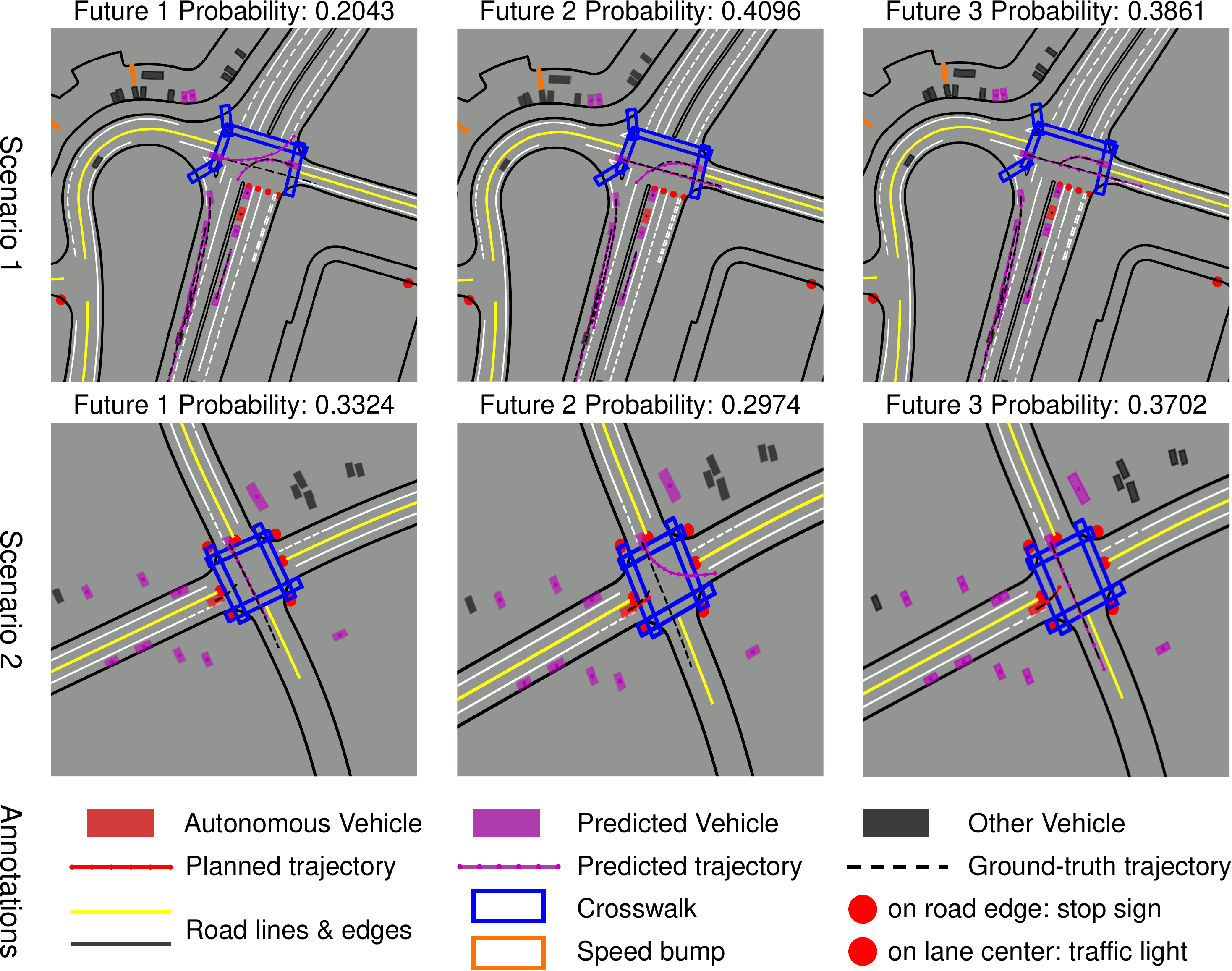}
    \caption{The multi-modal predictions given by the neural network predictor. The trajectories of the ten nearest agents to the AV are predicted by the neural network.}
    \label{fig4}
    \vspace{-0.31cm}
\end{figure*}

\textbf{Multi-modal joint prediction}. Fig. \ref{fig4} shows two representative scenarios where the neural network predictor makes multi-future predictions, as well as the probabilities of the predicted futures. In Scenario 1, the AV consistently chooses to stop behind the leading vehicle in all futures, while different interacting behaviors among the two agents in the intersection area are predicted. The future (joint trajectories) closest to ground truth is assigned with a higher probability. In Scenario 2, the predictor delivers different future predictions for both the AV and the other interacting agent at an unsignalized intersection. For example, the AV would choose to go if the other agent is predicted to turn left and stop if the other agent's trajectory is predicted to conflict with the AV's route. The close-to-ground truth futures are also assigned with higher probabilities. These results indicate that the neural network predictor is able to provide accurate multi-modal joint predictions for all agents in the scene and a good initial guess for the motion planner.

\begin{figure*}[htp]
    \centering
    \includegraphics[width=\linewidth]{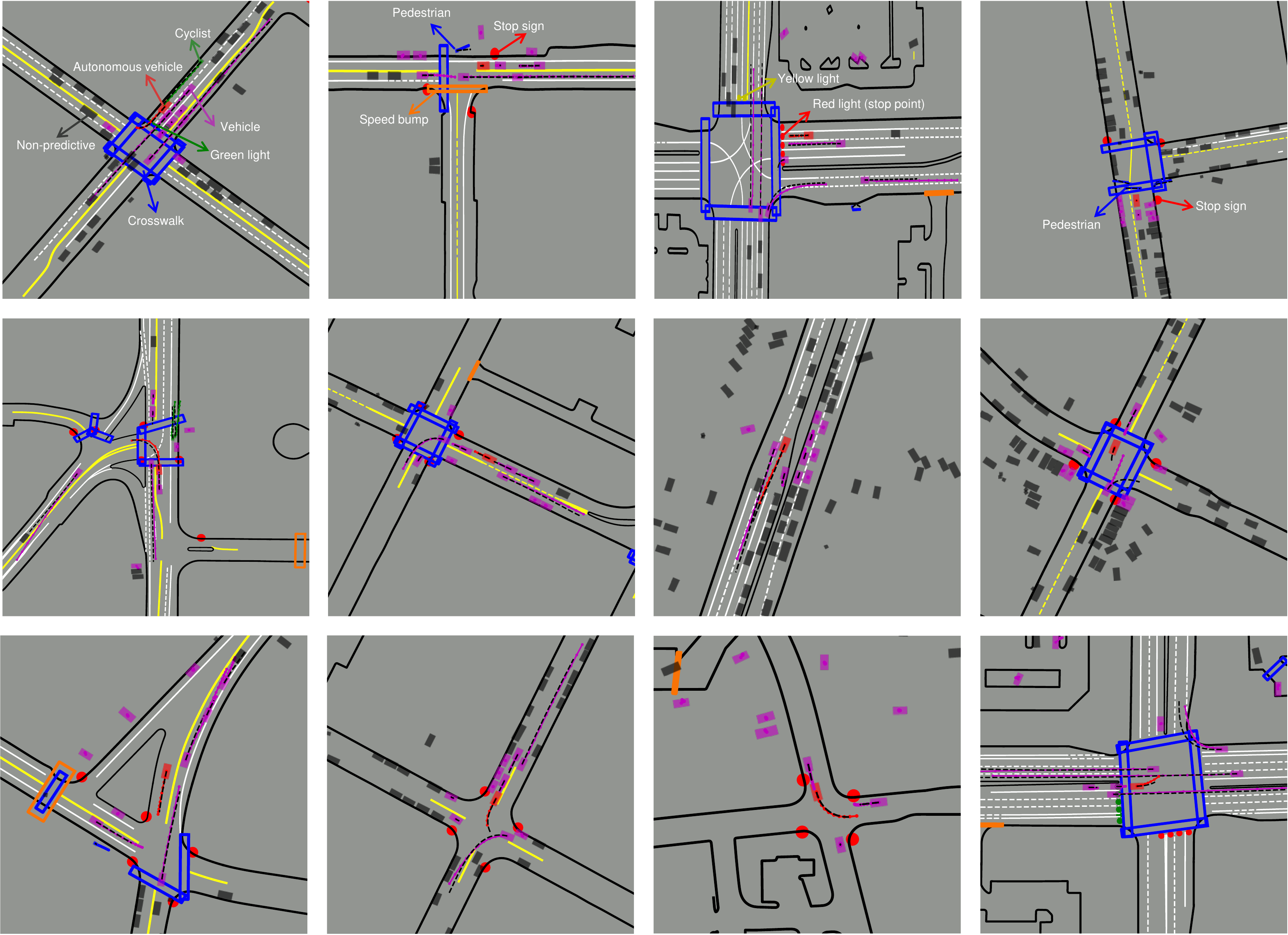}
    \caption{Qualitative results of the proposed framework in open-loop testing. The colored solid lines are the planned or predicted trajectories for AV or surrounding agents, and black dotted lines are the ground truth trajectories.}
    \label{fig5}
\end{figure*}

\begin{figure*}[htp]
    \centering
    \includegraphics[width=\linewidth]{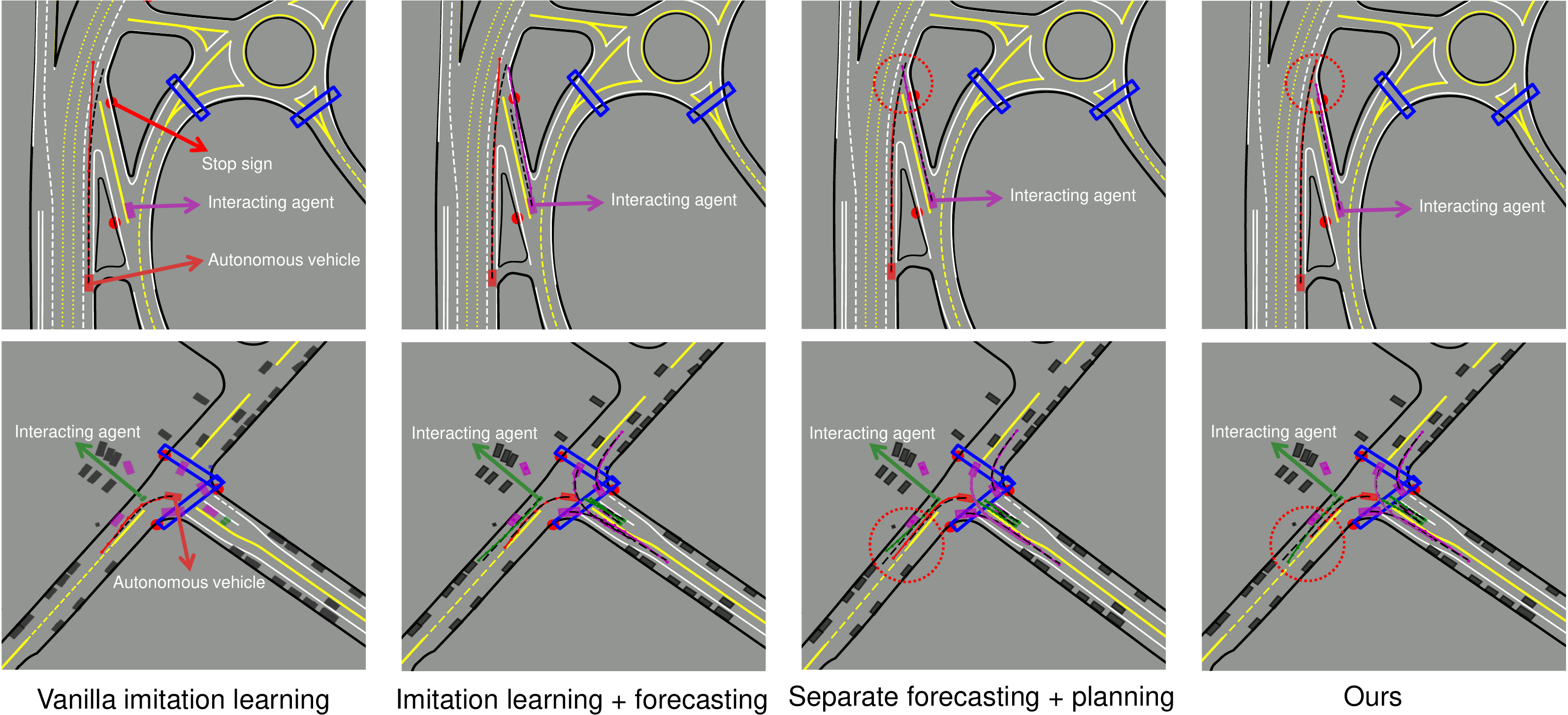}
    \caption{Qualitative comparison between the proposed method and baseline methods in open-loop testing. Top: interacting with a vehicle; bottom: interacting with a cyclist.}
    \label{fig6}
\end{figure*}

\textbf{Qualitative results}. Fig. \ref{fig5} displays some representative interactive scenarios, demonstrating the proposed framework's ability to plan a safe, traffic rule-compliant trajectory for the AV, based on the prediction results of other surrounding agents' future states (including vehicles, cyclists, and pedestrians). The results reveal that the motion planner is capable of handling a variety of urban driving scenarios involving a mixture of interacting traffic participants, including making a smooth right turn, yielding to pedestrians on the crosswalk, stopping at the red light, and yielding to another vehicle at an unprotected left turn, etc. In Fig. \ref{fig6}, we compare the proposed method with other baseline methods in two interactive scenarios. We can see that without explicit constraints, the neural network-planned trajectory gradually deviates from the lane centerline, while the motion planner's output trajectory adheres to the road structure. Our proposed method can generate planning-aware or planning-centric prediction results, which can help the planner deliver more close-to-human trajectories compared to using separate prediction results. For example, in the merging scenario, our method predicts that the interacting vehicle should yield to the AV at a stop sign, leading to more accurate prediction and planning results. In the interaction with a cyclist, our method predicts that the cyclist may move in front of the AV (however not really happen in the data) and the motion planner generates a slow-down trajectory to avoid potential risks, which is in accordance with the human-like trajectory. The result indicates that even if the prediction result is not entirely accurate, the planning module would benefit from such planning-centric results and produce human-like plans. The prediction of other agents in our framework is similar to the driver's belief rather than an exact reflection of the ground-truth dynamics.

\begin{table*}[htp]
\caption{Open-loop evaluation of the proposed method against baseline methods}
\centering
\begin{threeparttable}
\begin{tabular}{@{}l|ccc|ccc|ccc|cc@{}}
\toprule
\multirow{2}{*}{Method} &
\multicolumn{1}{c}{Collision} &
\multicolumn{1}{c}{Red light} &
\multicolumn{1}{c|}{Off route} &
\multicolumn{1}{c}{Acc.} &
\multicolumn{1}{c}{Jerk} &
\multicolumn{1}{c|}{Lat. Acc.} &
\multicolumn{3}{c|}{Planning error ($m$)} &
\multicolumn{2}{c}{Prediction error ($m$)} \\
                            &  (\%)            &  (\%)          &  (\%)             & ($m/s^2$)     & ($m/s^3$)     & ($m/s^2$)     & @1s            & @3s            & @5s              & ADE            & FDE \\ \midrule
IL                          & 5.469            & 2.772          &  4.816            & 0.546         & 2.881         & 3.151         & 0.175          & 1.416          & 4.194            & --             & --\\
IL+Prediction               & 3.930            & 1.670          &  4.542            & 0.672         & 4.004         & 3.939         & \textbf{0.127} & \textbf{0.892} & \textbf{2.901}   & 0.773          & 1.916\\
Sep. Plan+Pred.             & 1.813            & 1.327          &  0.527            & 0.269         & 0.149         & 0.077         & 0.238          & 1.251          & 3.466            & 0.766          & 1.908\\
DIPP (ours)                 &\textbf{1.802}    &\textbf{1.235}  & \textbf{0.506}    & 0.269         & 0.150         & 0.079         & 0.227          & 1.187          & 3.335            & \textbf{0.740} & \textbf{1.814}\\  \hline
Human                       & --               & --        &  --               & 0.620         & 1.707         & 0.101         & --             & --             & --               & --             & --\\ \bottomrule
\end{tabular}
\begin{tablenotes}
\footnotesize
\item Sep. Plan+Pred.: separated plan + prediction, Acc.: acceleration, Lat. Acc.: Lateral acceleration
\end{tablenotes}
\end{threeparttable}
\label{tab1}
\end{table*}

\begin{figure}[htp]
    \centering
    \includegraphics[width=0.95\linewidth]{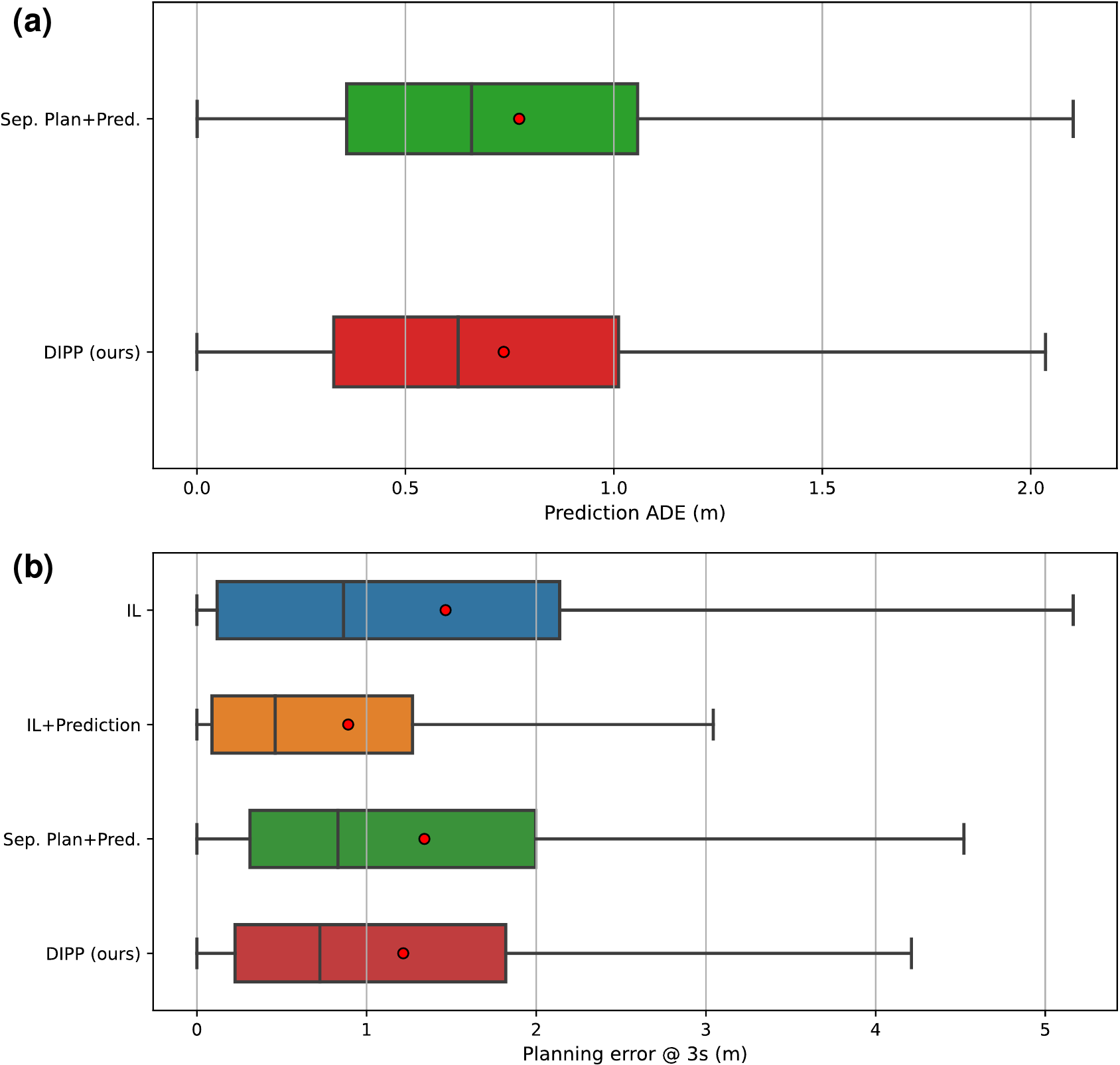}
    \caption{Boxplots of the prediction and planning errors of the proposed method and other baselines in the open-loop planning test: (a) prediction ADE, (b) planning error @3s.}
    \label{fig6.5}
\end{figure}

\textbf{Quantitative results}. Following the previously defined evaluation metrics, we list the quantitative results of our method and other baseline methods in Table \ref{tab1}. The results of open-loop testing indicate that the proposed method significantly outperforms the imitation learning-based methods in terms of collision and traffic rule violation rates. The imitation learning-based methods have a much higher off-route rate due to the lack of explicit constraints on following the route, and without considering the ride comfort, they yield an unacceptable lateral acceleration, which is significantly worse than the planning-based methods and human driving. On the other hand, the planning-based methods deliver much lower acceleration and jerk, which ensures smoothness and ride comfort. Because the neural network is only trained to predict the trajectories by direct regression, imitation learning with the multi-agent prediction sub-task method achieves the smallest planning error (compared with the ground-truth trajectories). However, since imitation learning methods ignore other important factors, they perform very poorly in the closed-loop test and barely finish a task (see Section \ref{r-closed}). One major conclusion from the results is that our proposed DIPP method, which jointly trains the predictor and planner, outperforms the planner with a separate trained prediction module, in terms of both planning and prediction errors. The improvements in DIPP may come from two possible ways: 1) the additional regularization from the planning module enables the prediction module to make fewer prediction errors that would negatively affect planning (e.g., large deviations from lane center and unnecessary stops), and 2) without significantly impacting raw prediction accuracy, the influence of the planning task enables the prediction module to make planning-centric results that would lead to better and human-like plans (e.g., react to possible cut-in of other agents). Another interesting finding is that though the vanilla imitation learning method performs the worst, adding the prediction sub-task can significantly reduce the collision rate and planning error, which shows the importance of multi-agent joint prediction. Fig. \ref{fig6.5} shows the boxplots of the prediction error of the proposed method compared against the separated prediction-planning method and the planning error of all methods in open-loop tests. The results further clarify the conclusion that differentiable and integrated training of the prediction and planning modules (DIPP) can improve both the prediction and planning metrics in open-loop tests.

\subsection{Closed-loop test}
\label{r-closed}

\begin{table*}[htp]
\caption{Closed-loop evaluation of the proposed method against baseline methods}
\centering
\begin{threeparttable}
\begin{tabular}{@{}l|ccc|ccc|ccc@{}}
\toprule
\multirow{2}{*}{Method} &
\multicolumn{1}{c}{Collision} &
\multicolumn{1}{c}{Off route} &
\multicolumn{1}{c|}{Progress} &
\multicolumn{1}{c}{Acc.} &
\multicolumn{1}{c}{Jerk} &
\multicolumn{1}{c|}{Lat. Acc.} &
\multicolumn{3}{c}{Position error ($m$)} \\
                            &  (\%)         &  (\%)      & $(m)$         & ($m/s^2$)     & ($m/s^3$)    & ($m/s^2$)     & @3s            & @5s            & @10s \\ \midrule
IL                          & 40            & 60         & 8.23          & 1.857         & 3.249        & 1.872         & 11.29          & 19.42          & 43.63  \\
IL+Prediction               & 42            & 58         & 11.05         & 1.345         & 1.977        & 0.813         & 11.03          & 19.13          & 42.25  \\
CQL                         & 50            & 50         & 16.38         & 3.456         & 8.798        & 2.154         & 12.85          & 20.18          & 45.10   \\
IDM                         & 28            & 0          & 49.44         & 0.885         & 2.236        & 0.081         & 3.895          & 9.568          & 30.32   \\
Sep. Plan+Pred.             & 7             & 0          & 76.28         & 0.638         & 1.313        & 0.058         & 1.869          & 4.182          & 10.17   \\
DIPP (Ours)                 &\textbf{5}     &\textbf{0}  &\textbf{77.57} & 0.624         & 1.209        & 0.069         &\textbf{1.726}  & \textbf{3.913} & \textbf{9.365}  \\ \hline
Human                       & --            & --         &  --           & 0.621         & 2.067        & 0.105         & --             & --             & --    \\ \bottomrule
\end{tabular}
\begin{tablenotes}
\footnotesize
\item Sep. Plan+Pred: separated plan + prediction, Acc.: acceleration, Lat. Acc.: Lateral acceleration
\end{tablenotes}
\end{threeparttable}
\label{tab2}
\end{table*}

\textbf{Qualitative results}. To fully reveal the planning performance of our proposed method, we conduct a closed-loop test with 100 replayed scenes from the testing set. Specifically, we build a log-replay simulator to roll out the AV's planned trajectory, where other agents follow their original tracks in the dataset and only the AV's state gets updated. At each timestep, the AV plans a trajectory according to the prediction result and takes the first action from its planned trajectory, and replans at the next time step. The simulation horizon is 15 seconds and the interval is 0.1 seconds. To evaluate the closed-loop testing performance, we add a progress metric, which measures the running distance of the AV until it reaches the end of the scene, collides with other agents, or goes off route. We measure the position error between the AV's rollout trajectory and its ground-truth one at several time steps to reflect the similarity to the human driver. The results of the closed-loop test are given in Table \ref{tab2} and we provide supplementary videos\footnote{\href{https://mczhi.github.io/DIPP/}{https://mczhi.github.io/DIPP/}} of our method navigating in a variety of urban scenarios for better visualization and evaluation of our method. It is important to note that the collision rate is just a lower bound of safety since other agents do not react to the AV (e.g., the ego vehicle runs slower than the human driver in the data). The supplementary videos encompass various driving scenes where the proposed framework can handle different tasks such as cruising, obeying traffic lights, making smooth turns, and more importantly interacting with other road users. We also add some scenarios where other vehicles would take adversarial actions toward the AV (not intentionally as the AV has deviated from its original track), but our planning framework can handle such emergencies and avoid collisions. 

\textbf{Qualitative results}. The results in Table \ref{tab2} indicate that the imitation learning-based methods cannot finish a single scene without causing a collision or going off route. Although the planning error of the IL+prediction method is the smallest in open-loop testing, the position errors of IL-based methods are significantly worse. The ride comfort is also compromised as IL-based methods deliver much higher jerk and acceleration. This is because the IL-based methods suffer from distributional shifts, which means the compounding errors lead the AV to a situation that deviates from the training distribution and degrade its performance. Similar to IL-based methods, the CQL method performs poorly in the test and cannot finish a single task because of the difficulty of training and labeling the correct reward for human driving behaviors. On the other hand, the planning-based methods leveraging structural information and domain knowledge perform significantly better despite learning purely from an offline dataset. The motion planner with a separate trained prediction module shows similar yet worse performance compared to our proposed method, which highlights the advantages of integrated training of the planning and prediction modules. Aside from its simplicity and few parameters, the IDM method is more effective in closed-loop testing since we only regulate the ego vehicle's longitudinal motion on the reference route. However, the performance of the IDM method is inferior to the planning-based approach, which is more flexible and can handle highly interactive and complex scenarios.

\subsection{Ablation study}
\label{abla}
\textbf{Effects of learnable components}.
To unveil the importance and function of each key component in our proposed framework, an ablation study is conducted. We set up three baselines in which the learnable components (i.e., initial plan, cost function, and prediction) are dropped out one at a time. We also set up an oracle method that uses the ground truth future trajectories of other agents as the prediction results to examine the capabilities of the proposed motion planner. For the non-learnable cost function baseline, the cost function used by the motion planner is manually tuned to balance the different cost terms. For the non-learnable initialization baseline, the initial guess to the motion planner is fixed as $(a_t=0, \delta_t=0)$ at all timesteps. For the non-learnable prediction baseline, the motion prediction module is a constant turn rate and velocity model. The results of the ablation study are summarized in Table \ref{tab3} and Table \ref{tab4}. In open-loop testing, we select the planning and prediction errors as the main metrics, and the planning error is averaged over three timesteps (1 s, 3 s, and 5 s) and ADE is used as the prediction error. In closed-loop testing, we employ the failure rate (sum of the collision and off-route rates), progress, and average position error (3 s, 5 s, and 10 s) as the main metrics. 

\begin{table}[htp]
\caption{Ablation study on the importance of each component in open-loop testing}
\centering
\begin{tabular}{@{}l|cc@{}}
\toprule
Model                       & Planning error (m) & Prediction error (m)    \\ \midrule
Base (ours)                 &  1.583             & 0.740   \\
Base w/ oracle prediction   &  1.388             & -- \\
No learnable cost function  &  1.834             & 0.740  \\
No learnable initialization &  2.672             & 0.740    \\
No learnable prediction     &  1.811             & 1.871   \\ \bottomrule
\end{tabular}%
\label{tab3}
\end{table}

\begin{table}[htp]
\caption{Ablation study on the importance of each component in closed-loop testing}
\centering
\resizebox{\linewidth}{!}{%
\begin{tabular}{@{}l|cccc@{}}
\toprule
Model                       & Failure (\%)  & Progress (m)  & Position error (m)   \\ \midrule
Base (ours)                 &  5            & 76.28         & 4.083   \\
Base w/ oracle prediction   &  6            & 76.16         & 4.153    \\
No learnable cost function  &  13           & 69.09         & 5.881  \\
No learnable initialization &  17           & 57.52         & 6.571  \\
No learnable prediction     &  19           & 44.01         & 9.971  \\ \bottomrule
\end{tabular}%
}
\label{tab4}
\end{table}

The open-loop testing results in Table \ref{tab3} indicate that learnable initialization is the most important part of the framework to ensure the planned trajectory is close to the human-driving one. This suggests that a good initial plan for the planner is critical to solving the optimization and learnable initialization could substantially improve the planning performance. Another important factor for planning is the cost function, and we demonstrate that learning the cost function from data can render the planned trajectories closer to humans. Surprisingly, using only a physics-based non-learnable prediction model would not significantly deteriorate the planning performance in an open-loop setting though the prediction error is notably higher. Using the oracle predictor can improve the motion planner's performance in open-loop testing, which suggests that the proposed motion planner performs well and could achieve even better performance with more training data and improved prediction accuracy. The closed-loop testing results in Table \ref{tab4} reveal that the prediction module plays a more critical role in ensuring safety and human likeness. The gap between open-loop and closed-loop performance is due to that the closed-loop testing set contains many interactive scenarios, where accurately predicting other agents' future states is required. In contrast to open-loop testing, our method can deliver comparable performance to the oracle prediction method in closed-loop testing, marking the advantage of our method (planning-aware prediction) when deploying the system in practice. In addition, learnable initialization and cost function are necessary to stabilize the planning process (solving the optimization problem) and obtain better planning performance, which otherwise could lead to instability to find viable solutions and consequently worse safety and reliability. In summary, all three learnable components in our framework are integral to maintaining the final planning performance.

\textbf{Effects of weights of the training loss function}.
We investigate the influence of the weights in the loss function (Eq. \ref{loss}) and test the trained models in the open-loop planning setting. To reveal the sensitivity of the framework performance to these values, we set up the following groups of loss function weights: (1) the default parameter setting in the experiment, (2) the weight for the prediction loss $\lambda_1$ is increased by $5\times$, (3) the weight for the prediction loss $\lambda_1$ is decreased to $1/5$ of the default value, (4) the weight for the score loss $\lambda_2$ is increased by $5\times$. The weights for imitation loss $\lambda_3$ and planning cost $\lambda_4$ are fixed, and the same metrics in Table \ref{tab3} are used to evaluate the framework's performance.

\begin{table}[htp]
\caption{Effects of weights of the loss function in open-loop testing}
\centering
\begin{tabular}{@{}lccc|cc@{}}
\toprule
No. & $\lambda_1$ & $\lambda_2$  & $\lambda_3$   & Planning error (m)    & Prediction error (m) \\ \midrule
1   & 0.5         & 1            & 1             &  1.583                & 0.740  \\
2   & 2.5         & 1            & 1             &  1.837                & 0.733   \\
3   & 0.1         & 1            & 1             &  1.642                & 1.244   \\
4   & 0.5         & 5            & 1             &  1.704                & 0.820   \\  \bottomrule
\end{tabular}
\label{tab5}
\end{table}

The results in Table \ref{tab5} indicate that the default loss parameters in our experiments (No. 1) achieve the best planning performance. Although increasing the weight of prediction loss (No. 2) can slightly improve the prediction error, the planning performance is compromised because the loss signal of the initial plan or cost function in motion planning is largely neglected in the total loss function. However, decreasing the weight of prediction loss (No. 3) still does not bring better planning performance because the prediction results are required to be more accurate for both the ego vehicle and other agents. In addition, increasing the weight of score loss (No. 4) may not improve probability evaluation and even lead to worse planning or prediction performance. The results suggest that our multi-task learning framework requires a trade-off between the planning and prediction tasks, and the default loss function weights can strike a better balance between the tasks.

\textbf{Effects of the step size and planning iterations}.
We investigate the effects of the step size ($\alpha$ in Eq. \ref{update}) and planning iterations in the motion planner on the final open-loop planning test. Note that the different parameters are only applied in the training process while the parameters of the motion planner in the testing process are the same. We explore different parameter settings in Table \ref{tab6}, where the value of the step size $\alpha$ ranges from 0.2 to 1, and the maximum planning iterations is 3 for speed consideration. The same metrics for open-loop testing are used to evaluate the performance of the trained models.

\begin{table}[htp]
\caption{Effects of the planner update step size and planning iterations in open-loop testing}
\centering
\resizebox{\linewidth}{!}{%
\begin{tabular}{@{}cc|cc@{}}
\toprule
step size $\alpha$  & planning iteration    & Planning error (m)    & Prediction error (m) \\ \midrule
0.4                 & 2                     &  1.583                & 0.740  \\
1.0                 & 2                     &  1.856                & 0.867   \\
0.4                 & 3                     &  1.551                & 0.725   \\
0.2                 & 3                     &  1.549                & 0.797  \\  \bottomrule
\end{tabular}
}
\label{tab6}
\end{table}

The results in Table \ref{tab6} indicate that using the default planner setting provides satisfactory results in the final open-loop planning test. Setting the step size to the maximum value ($\alpha=1$) could significantly impair the testing performance because the large step size makes the optimization process unstable, which negatively impacts the learning process of the prediction module and the cost function weights. While increasing the number of planning iterations with the default step size ($\alpha=0.4$) could slightly improve the planning and prediction metrics, this would also result in longer computing time for the planner, making it an unnecessary choice. On the other hand, using a smaller step size ($\alpha=0.2$) along with additional planning iterations could also achieve comparable performance to the default setting, but may still be an unnecessary choice. Overall, the default planner parameter setting provides a good balance between final performance and computational efficiency.

\begin{table*}[htp]
\caption{Detailed results for the learned planning cost}
\centering
\resizebox{\linewidth}{!}{%
\begin{tabular}{@{}l|ccccccccc|cc@{}}
\toprule
Cost Weights    & $\omega_1$ ($\mathbf{c}^{speed}$) & $\omega_2$ ($\mathbf{c}^{acc}$) & $\omega_3$ ($\mathbf{c}^{jerk}$) & $\omega_4$ ($\mathbf{c}^{steer}$) & $\omega_5$ ($\mathbf{c}^{rate}$) & $\omega_6$ ($\mathbf{c}^{pos}$) & $\omega_7$ ($\mathbf{c}^{head}$) &  $\omega_8$ ($\mathbf{c}^{traffic}$) & $\omega_9$ ($\mathbf{c}^{safety}$) & Planning error (m) &  Failure (\%)\\ \midrule
Hand-tuned      &  0.1    & 0.5    & 0.1     & 0.01     &  0.5     & 0.5    &  5     &  10    & 10    & 1.834          & 13\\
Learned         &  0.055  & 0.29   & 0.13    & 0.014    &  0.46    & 0.31   &  3.23  &  10    & 10    & \textbf{1.583} & \textbf{5} \\ \bottomrule
\end{tabular}%
}
\label{tab7}
\end{table*}

\textbf{Results of cost function learning}.
We report detailed results for the learned cost function and hand-tuned one in Table \ref{tab7}. The traffic light violation and safety weights are fixed at $10$ because they are considered constraints, and other parameters are learnable. The results indicate that using the learned cost function as opposed to the hand-tuned cost function, significantly reduces planning errors in the open-loop planning test and the failure rate in the closed-loop test. Generally, the learned weights are reasonable from the viewpoint of human drivers. Adhering to the driving route (for both positions and directions) is the most critical factor, and driving comfort (acceleration and change rate of steering) is also an important factor to optimize. The steering angle is not a significant factor in the cost since the vehicle primarily needs to follow the route, but speed and jerk cost terms are assigned with smaller weights due to their relatively large scales. It is important to note that the cost terms have different scales, and the weights are learned to balance the scales to facilitate the optimization process. This means that the learned cost function weights may not reflect their actual weights in human driving and are thereby unsuitable for other frameworks.

\subsection{Discussions}
\textbf{Strengths}. The direct benefit of our method is to integrate structural information and domain knowledge into machine learning models, or from a different perspective, to apply machine learning models to some hard-to-solve points in traditional optimization problems. We apply our method to the challenging autonomous driving task and focus on solving real-world problems (e.g., how to specify the cost function and how other agents' actions would affect the ego) by learning from real-world data. The integration of prediction, cost function, and planning modules in our framework leads to end-to-end, decision-focused learning, that trains the whole pipeline to optimize the final planning performance directly. Experiments demonstrate that the proposed method has excellent open-loop and closed-loop performance and robustness against distributional shifts and adversarial agents, despite only learning from offline data. In addition, the proposed method outperforms the separated planning and prediction method in both planning and prediction performance, which underscores the advantage of our method to integrate the prediction and planning modules in an end-to-end differentiable way.

\textbf{Limitations and future work}. Some limitations of this work should be acknowledged. First is the computation time of the framework. The average runtime of the prediction+planning step is 0.012 s per sample at training (with NVIDIA RTX 3080 GPU, 2 iterations) and 1.78 s at testing (with AMD 3900X CPU, max 50 iterations). The bottleneck in the framework is the computation of Jacobin in solving the least-squares optimization. However, we need to note that this work is just a prototype of the proposed framework and the running efficiency has not been optimized. We plan to further improve the efficiency of the computation pipeline to reduce the runtime in solving the optimization, potentially enabling the framework to run near real-time. Another limitation is that we do not thoroughly consider the uncertainty or multi-modality of human behavior prediction, only taking the most probable one as the planner input. In future work, the planner can take all modalities and uncertainties along the timesteps into account to make more informed decisions or perform contingency planning.

\section{Conclusions}
We propose a novel differentiable integrated multi-agent interactive prediction and motion planning framework, which is trained end-to-end from real-world driving data. A Transformer-based predictor is established to predict joint future trajectories of surrounding agents and provide an initial guess for the planner. The predicted trajectories, initial plan, and learnable cost function are channeled to an optimization-based differentiable motion planner, allowing every component in the framework to be differentiable. The framework is validated on a large-scale urban driving dataset in both open-loop and closed-loop testing. The open-loop testing results reveal that our proposed method outperforms traditional pipeline methods and IL-based methods in terms of planning and prediction performance, ride comfort, and safety metrics. The closed-loop testing results indicate that planning-based methods significantly outperform IL-based methods (though have better similarity to human trajectories in open-loop testing), which suggests that our method overcomes the distributional shift problem common in offline learning. Moreover, the proposed method outperforms the pipeline method in closed-loop testing, emphasizing the benefit of joint training of prediction and planning modules. In the ablation study, we find that all learnable components in the framework are integral to maintaining the stability of the optimizer and desired planning performance. The proposed framework has the potential to accelerate the development of autonomous driving decision-making systems by enabling all components to be learned from real-world data.



%





\ifCLASSOPTIONcaptionsoff
  \newpage
\fi





\bibliographystyle{IEEEtran}
\bibliography{egbib}

\vfill


\end{document}